\documentclass[conference]{IEEEtran}
\IEEEoverridecommandlockouts
\usepackage{booktabs}
\usepackage{hhline}
\usepackage{arydshln}
\usepackage{cite}
\usepackage{amsmath,amssymb,amsfonts}
\usepackage{algorithmic}
\usepackage{graphicx}
\usepackage{textcomp}
\usepackage{xcolor}
\def\BibTeX{{\rm B\kern-.05em{\sc i\kern-.025em b}\kern-.08em
		T\kern-.1667em\lower.7ex\hbox{E}\kern-.125emX}}

\usepackage[mathlines,switch]{lineno}
\usepackage{fancyhdr}

\begin{document}
	
	\title{Leveraging Swin Transformer for Local-to-Global Weakly Supervised Semantic Segmentation\\
	}
	
	\author{
		\IEEEauthorblockN{Rozhan Ahmadi}
		\IEEEauthorblockA{\textit{Department of Computer Engineering} \\
			\textit{Sharif University of Technology}\\
			Tehran, Iran \\
			roz.ahmadi@sharif.edu}
		\and
		\IEEEauthorblockN{Shohreh Kasaei}
		\IEEEauthorblockA{\textit{Department of Computer Engineering} \\
			\textit{Sharif University of Technology}\\
			Tehran, Iran \\
			kasaei@sharif.edu}
	}

	\maketitle
	
	\thispagestyle{plain}
	\fancypagestyle{plain}{
		\fancyhf{} 
		\fancyfoot[L]{979-8-3503-5049-4/24/\$31.00~\copyright2024~IEEE} 
		\fancyhead[L]{2024 13th Iranian/3rd International Machine Vision and Image Processing Conference (MVIP)}
		\renewcommand{\headrulewidth}{0pt}
		\renewcommand{\footrulewidth}{0pt}
	}

	\begin{abstract}
		In recent years, weakly supervised semantic segmentation using image-level labels as supervision has received significant attention in the field of computer vision. 
		Most existing methods have addressed the challenges arising from the lack of spatial information in these labels by focusing on facilitating supervised learning 
		through the generation of pseudo-labels from class activation maps (CAMs). Due to the localized pattern detection of Convolutional Neural Networks (CNNs), CAMs often 
		emphasize only the most discriminative parts of an object, making it challenging to accurately distinguish foreground objects from each other and the background. 
		Recent studies have shown that Vision Transformer (ViT) features, due to their global view, are more effective in capturing the scene layout than CNNs. However, 
		the use of hierarchical ViTs has not been extensively explored in this field. This work explores the use of Swin Transformer by proposing "SWTformer" to enhance 
		the accuracy of the initial seed CAMs by bringing local and global views together. 
		SWTformer-V1 generates class probabilities and CAMs using only the patch tokens as features. SWTformer-V2 incorporates a multi-scale feature fusion mechanism to 
		extract additional information and utilizes a background-aware mechanism to generate more accurate localization maps with improved cross-object discrimination. 
		Based on experiments on the PascalVOC 2012 dataset, SWTformer-V1 achieves a 0.98\% mAP higher localization accuracy, outperforming state-of-the-art models. 
		It also yields comparable performance by 0.82\% mIoU on average higher than other methods in generating initial localization maps, depending only on the 
		classification network. SWTformer-V2 further improves the accuracy of the generated seed CAMs by 5.32\% mIoU, further proving the effectiveness of the 
		local-to-global view provided by the Swin transformer. Code available at: https://github.com/RozhanAhmadi/SWTformer
	\end{abstract}
	
	
	\section{Introduction}
	Semantic segmentation is a crucial task in computer vision, where a deep learning model is used to classify every pixel 
	in an image.   The progress in fully supervised learning has led to significant results in semantic segmentation, leading 
	to its employment in various applications such as medical imaging \cite{b1} and autonomous driving \cite{b2}. However, due 
	to the massive amount of manual and time-consuming pixel-level image annotations, the cost of obtaining data for this task 
	can be substantial.
	
	\begin{figure}[t]
		\centering
		\includegraphics[width=\columnwidth]{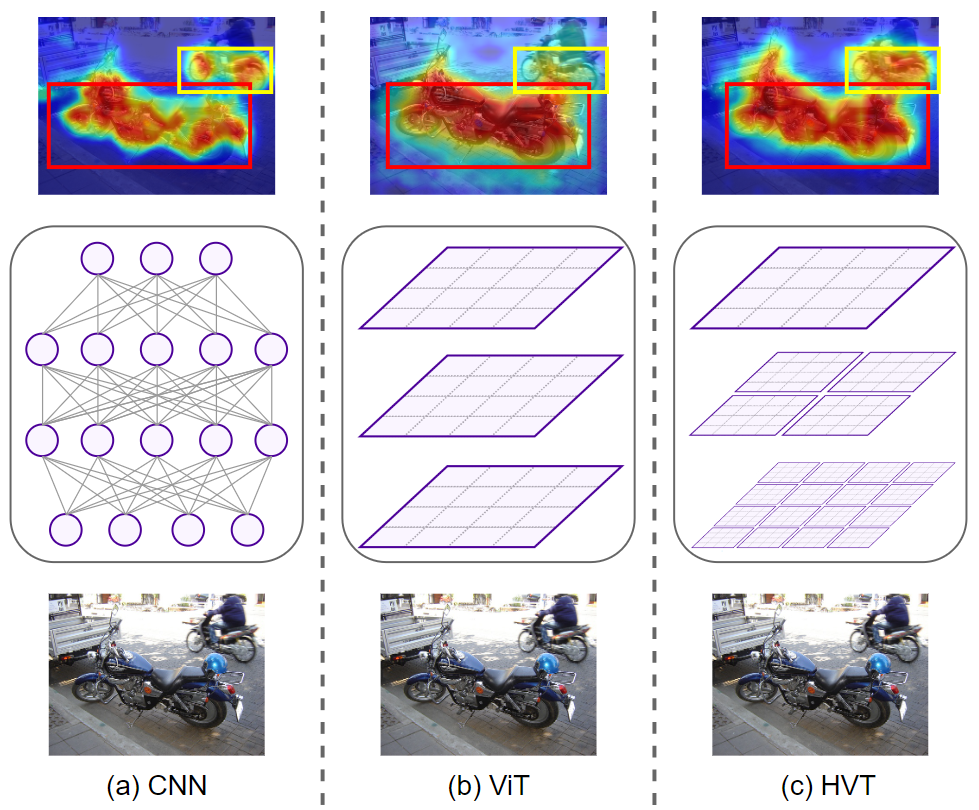}
		\caption{Class activation maps generated by a (a) CNN (Resnet-50), (b) ViT (DeiT-S) and (c) HVT (Swin-T). Red and yellow boxes indicate the large and small scale objects relative to the image size.}
		\label{fig_1}
	\end{figure}
	
	In recent years, to reduce annotation costs, many studies have gravitated towards weakly supervised semantic segmentation (WSSS). This approach utilizes weak labels including bounding boxes \cite{b3}, scribbles \cite{b4}, points \cite{b5}, and image-level labels to train the segmentation model. Among these, image-level labels are the easiest to annotate, making them the primary focus of many studies. Semantic segmentation using only image-level labels is a complex task due to the lack of spatial information regarding object distributions. To tackle this, many approaches use a three-step pipeline. Initially, seed localization maps are generated using class activation maps (CAMs) \cite{b6} from an image classification network. These seed maps only focus on discriminatory parts of objects and have activation limitations. Subsequently, the seed maps are refined to cover object areas more comprehensively and then used to generate pseudo-labels. These pseudo-labels are then used to train a fully supervised semantic segmentation network. The flow of this pipeline shows that the success of WSSS heavily relies on the quality of the initial CAMs. Therefore, significant research efforts have been dedicated to enhancing the accuracy of these initial maps.
	
	Convolutional neural networks (CNNs) are the most commonly integrated backbone models for WSSS. Despite the progress in this field, the proposed strategies still face limitations in generating complete object regions, caused by convolutions perceiving only small-range feature relations and only providing local reception of a scene, Fig.~\ref{fig_1} (a). On the other hand, the utilization of Vision Transformers (ViTs) is a relatively recent area of research in WSSS. Unlike CNNs, vision transformers are capable of capturing long-range semantic dependencies and provide a global understanding of the scene. However, replacing CNNs with ViTs results in a trade-off between gaining more complete regions for large-scale objects and losing fine-grained details, Fig.~\ref{fig_1} (b). Hierarchical Vision Transformer (HVTs) is a type of ViT designed to integrate the strengths of both CNNs and ViTs.  Unlike common ViTs, which produce feature maps of a single low resolution, HVTs use a hierarchical design to generate feature maps at multiple resolutions. This approach allows HVTs to effectively capture both local and global contextual information, Fig.~\ref{fig_1} (c), making them well-suited for accurate multi-scale object localization. Despite HVTs' promising performance in various tasks, their application in WSSS has yet to be explored.
	
	To investigate the validity of this idea, this work proposes a novel approach, termed SWTformer.   SWTformer-V1 is designed to leverage the state-of-the-art HVT, Swin Transformer \cite{b7}, as the backbone classifier network supervised by image-level labels. This is particularly challenging as Swin uses only patch tokens, while common ViTs used in WSSS mostly depend on class tokens. The shifted window mechanism of Swin also requires careful consideration in parameter adjustment and fine-tuning when used as a backbone for WSSS. Additionally, the success in the utilization of common ViTs is mainly attributed to Attention Roll-Out \cite{b8}, a mechanism that accumulates attention maps from the ViT layers and allows for a more specific interpretation of attention flow in the network. However, due to the shifted windows mechanism and hierarchical multi-scale features of Swin, this approach is not yet applicable to this network.  To overcome these limitations and extract further contextual information, SWTformer-V2 proposes a multi-scale feature fusion module and employs it in a background-aware refinement mechanism to generate more comprehensive localization masks with higher cross-object discrimination accuracy.
	
	The main contributions of this work can be summarized as:
	\begin{itemize}
		\item Proposing the first hierarchical transformer-based solution (SWTformer) for generating initial CAMs in WSSS, to address the trade-off between limitations of CNN’s local receptive field and ViT’s global view of a scene.
		\item SWTformer-V1 proposes an approach to leverage the Swin Transformer as a backbone for classification and initial CAM generation using only patch tokens.
		\item SWTformer-V2 addresses the in-feasibility of performing Attention Roll-Out on the Swin Transformer architecture and proposes a solution in the form of utilizing hierarchical feature fusion and background-aware refinement mechanism.
		\item Validating the effectiveness of the proposed methods with extensive experiments on the PascalVOC 2012 dataset.
	\end{itemize}
	
	\begin{figure*}[htbp]
	\centering
	\includegraphics[width=\textwidth]{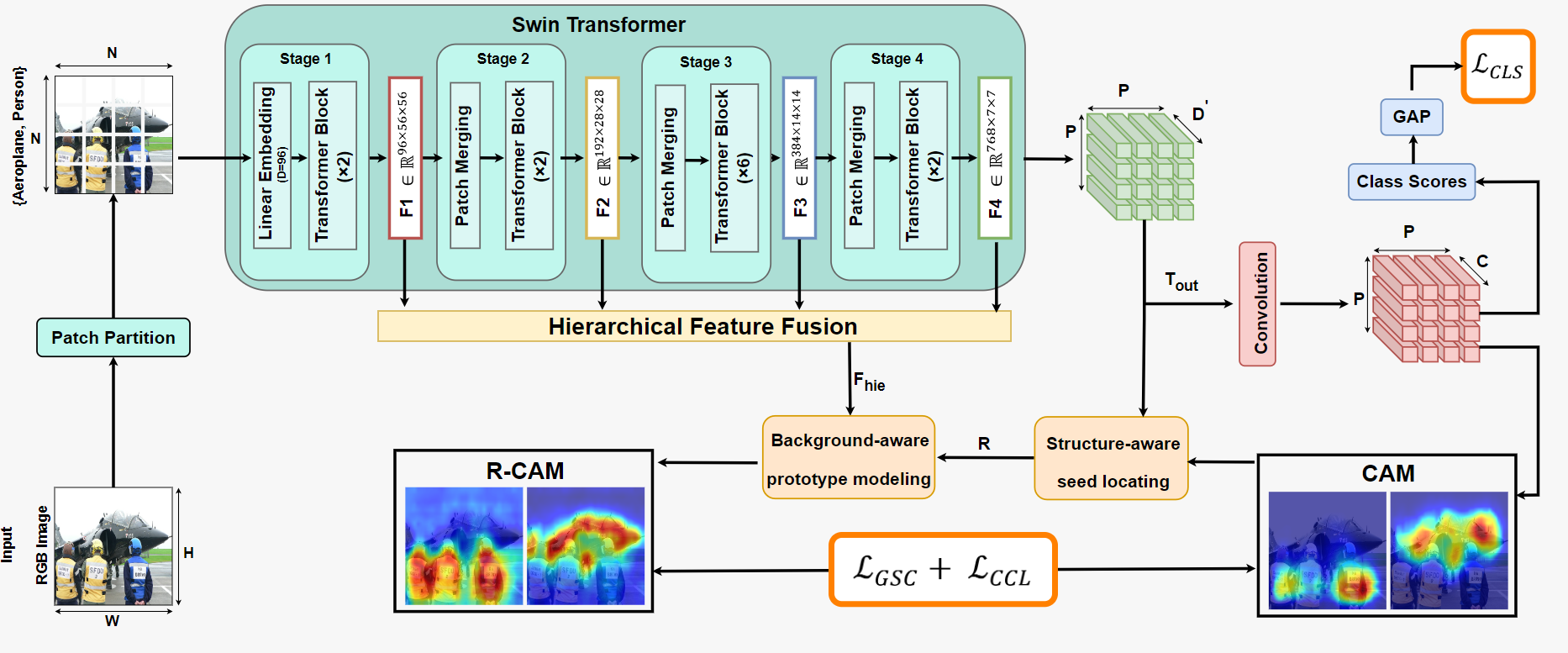}
	\caption{An overview of the proposed SWTformer (V2). The backbone is the Swin-T version of the Swin Transformer and the training of the model is optimized by the CLS, GSC and CCL loss functions. The “Structure-aware seed locating” and “Background-aware prototype modeling” modules are adopted from SIPE \cite{b23} with modifications.}
	\label{fig_2}
	\end{figure*}

	\section{Related Work}
	
	\subsection{Vision Transformers}
	
	In recent years, Vision Transformers (ViTs) have significantly revolutionized the field of computer vision. ViT \cite{b9} is a deep learning model that transforms an input image into a sequence of tokens called patch tokens, adds a class token to serve as a representation of the entire image and then analyzes the visual data by applying several layers of multi-head self-attention blocks to these tokens. This self-attention mechanism allows ViT to capture global information and long-range dependencies in the data.  Built on ViT, DeiT \cite{b10} proposes new data augmentation methods and introduces a distillation token. Although ViTs succeed in capturing global context, they have limitations in capturing local details. To solve this, Conformer \cite{b12} combines a CNN branch with a ViT branch. However, combining two complex networks requires significant training adjustments and computation effort. To this end, Hierarchical Vision Transformers (HVTs) serve as the most effective solutions for combining the strengths of ViTs and CNNs by starting from fine-grained local details and moving up to long-range global dependencies. For instance, T2T \cite{b13} proposes a token-to-token transformation module that gradually merges neighboring patch tokens.  PVT \cite{b4} reduces the token sequence length by merging tokens along a pyramid structure. Swin Transformer \cite{b7} utilizes a novel patch merging module and a shifted window self-attention mechanism. This approach allows smaller groups of patches to be mixed together, enabling the model to capture long-range feature dependencies more accurately.
	
	\subsection{Weakly Supervised Semantic Segmentation with CNNs}
	
	Recent studies on WSSS mostly use image-level labels for supervision and rely on Class Activation Maps (CAM) \cite{b6} to localize objects. These works utilize a CNN as the classification backbone to generate the seed localization maps and then train a fully supervised segmentation model, using the seed maps as pseudo labels. CAM is calculated for each class through a weighted combination of the feature maps in the last layer of a CNN, the weights being the last fully connected layer’s weights belonging to the target class. CAMs help to visualize the most discriminative regions of an image influencing a classifier's decision-making. However, CAMs have limitations in comprehensively activating object regions and accurately distinguishing foreground objects from each other and from the background. A common approach to solve this is adding a post-processing stage prior to segmentation. PSA\cite{b15} and IRN\cite{b16} achieve this by refining the initial seed maps into pseudo masks through iterative seed region growing methods.  Additionally, as the performance of WSSS is heavily dependent on the quality of the initial CAMs, various methods have been suggested to create more accurate initial activation maps. These methods include adversarial erasing \cite{b17}, \cite{b18}, cross-affinity extraction modules and contrastive learning \cite{b19}, \cite{b20}, attention mechanism \cite{b21}, \cite{b22} and self-supervised learning \cite{b23}, \cite{b24}. 

	\subsection{Weakly Supervised Semantic Segmentation with ViTs}
	
	With Vision Transformers (ViTs) making significant progress in various tasks, some recent works have utilized them for WSSS. AFA \cite{b25} proposes refining initial pseudo labels using global semantic affinity learned from self-attention. MCTformer \cite{b26} replaces ViT’s singular class token with multiple tokens, each corresponding to a particular semantic class. It also employs patch affinity learned from attention maps to refine the initial CAMs. ViT-PCM \cite{b27} proposes an end-to-end CAM independent framework relying on ViT’s spatial characteristics. ToCo \cite{b28} addresses the over-smoothing issues of ViTs by using the model’s intermediate knowledge to supervise its output features. TransCam \cite{b29} adopts Conformer \cite{b12} by proposing to use the attention weights of the ViT branch to refine the CAMs generated from the CNN branch. It is worth mentioning that these studies mostly rely on class tokens, inspired by observations made in DINO \cite{b30} that class tokens and their attention to patch tokens contain useful knowledge regarding the semantic layout of a scene. Hierarchical Vision Transformers are a rather recent development in the field of vision transformers and have not yet been introduced to WSSS. HVTs are expected to capture scene layout more effectively than CNNs and ViTs. However, their specific impact in the context of WSSS remains an open area for research.
	
	\section{Proposed Method}
	\subsection{Overview}
	This paper introduces SWTformer, illustrated in Fig.~\ref{fig_2}, a novel framework that utilizes the Swin Transformer as the classifier backbone to generate initial localization maps for WSSS. Unlike traditional ViTs, Swin Transformer operates based on only patch tokens without the use of any class tokens. Inspired by \cite{b26}, SWTformer-V1 integrates a CAM module to use Swin’s output patch tokens for generating initial activation seeds and producing class scores to train the classifier. Due to Swin Transformer’s hierarchical structure, combining attention maps from intermediate layers is not feasible and is an active area of research. To overcome this limitation, SWTformer-V2 proposes the use of Swin’s multi-scale contextual information through a Hierarchical Feature Fusion (HFF) module instead of combining attention maps to learn semantic patch affinity. SWTformer-V2 builds the HFF module on SWTformer-V1 by employing it in a background-aware prototype exploration mechanism based on SIPE \cite{b23}. The primary goal of SWTformer-V2 is to refine the initially generated CAMs from SWTformer-V1, enabling the model to create more comprehensive object regions and accurately distinguish foreground objects from the background.
	
	\subsection{Generating Class Activation Maps from Patch Tokens}
	Given an image $ I\in \mathbb{R}^{3\times H \times W} $, the Swin encoder first partitions it into $ N\times N $ patches. The linear embedding module in the first transformer 
	block then projects these patches into a sequence of patch tokens $ T\in \mathbb{R}^{D\times N \times N} $ with $ D $ being the embedding dimension. A patch merging module 
	connects the three upcoming transformer blocks by doubling the embedding dimension of the input and downsizing it to half its size. As a result, the 
	encoder’s output token sequence can be represented as  $ T_{out}\in \mathbb{R}^{D^{'}\times P \times P} $, where $ D^{'}=8D $ and$  P=N/8 $. To generate class activation maps for $ C $ classes, a convolutional 
	layer with $ C $ output channels is applied to $ T_{out} $, converting it to a 2D feature map $ F_{out}\in \mathbb{R}^{C\times P \times P}$. $ F_{out} $ may contain negative values that are not meaningful for interpreting 
	the category distribution of patch tokens. In order to make the values of  $ F_{out}$ represent class probabilities, a ReLU function is applied to $ F_{out}$, followed by a feature normalization 
	function. This process results in feature maps $ C_{out}\in \mathbb{R}^{C\times P \times P}$  that are then upsampled to the size of the original image, producing the initial class activation maps $ M\in \mathbb{R}^{C\times H \times W}$.
	
	\subsection{Multi-label Classification Training}
	To perform the multi-label classification task, class scores $s\in \mathbb{R}^{C}$ for $C$ semantic categories are calculated by applying global average pooling to $F_{out}$, where $s_{c}$ denotes the predicted probability of the input belonging to class $c\in \{1, …, C\}$. A multi-label soft margin loss is then computed between each predicted class score $s_{c}$ and its corresponding ground-truth $\hat{s}_c$. The classification loss is then calculated by averaging over all $C$ classes,
	\begin{equation}
		\mathcal{L}_{CLS}=\frac{1}{C} \sum_{i=1}^C \hat{s}_c \log \left(s_{c}\right)+\left(1-\hat{s}_c\right) \log \left(1-s_{c}\right).
	\end{equation}
	
	\subsection{Hierarchical Feature Fusion}
\begin{figure}[ht]
		\centering
		\includegraphics[width=\columnwidth]{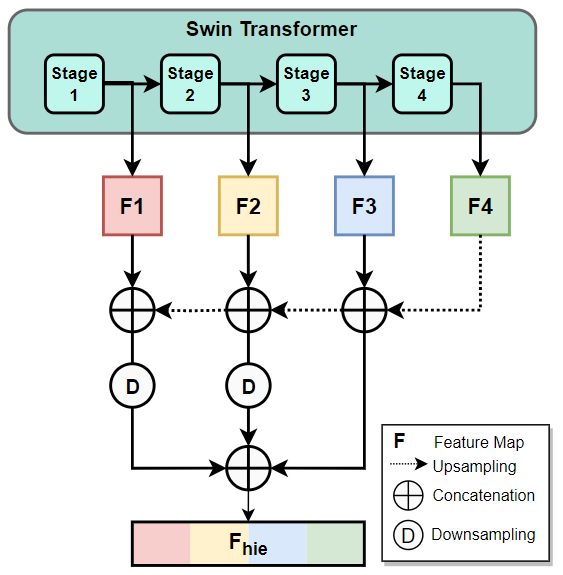}
		\caption{Illustration of the proposed hierarchical feature fusion (HFF) module in SWTformer.}
		\label{fig_3}
	\end{figure}

	The utilization of Swin's output patch tokens to calculate class scores and generate initial CAMs has been demonstrated in SWTformer-V1. In the field of deep learning, leveraging feature maps from both the final and intermediate layers of a hierarchical network is a well-established strategy \cite{b31}, \cite{b32}. Inspired by this, the suggested approach takes advantage of the distinct characteristics of information captured at different stages of the network. Shallow layers, which are closer to the input data, are capable of identifying low-level granular local features such as edges, texture and color. On the other hand, deep layers, as the network hierarchy is ascended, are capable of recognizing more abstract, high-level features and complex patterns. By fusing feature maps from both shallow and deep layers, the model can harness a comprehensive range of information, from simple to complex patterns. Using the model’s intermediate features during training won't cause any significant changes in computational cost, as these features have already been calculated in the model and can be directly used without much overhead. Fig.~\ref{fig_3} illustrates the proposed hierarchical feature fusion method. The HFF module extracts the output patch features from all four transformer blocks and concatenates them in two stages to maximize the semantic knowledge obtained. The upsampling in this module is achieved through bilinear interpolation, while the downsampling is performed using a convolutional layer. This module is specifically designed to be compatible with the Swin Transformer and outputs a new feature map $F_{hie}$ that contains the combined local to global semantic contexts of the scene.
	
	\subsection{Background-aware Prototype Exploration}
	In order to leverage the semantic knowledge encapsulated in  $F_{hie}$ for enhancing the initially generated CAMs and generating masks with more comprehensive object activation, SWTformer-V2 adapts and modifies the utilization strategy demonstrated by SIPE \cite{b23}, an architecture based on a CNN.  Given $M\in \mathbb{R}^{C\times H\times W}$ for $C$ foreground classes, to enhance the model's awareness of the background, an activation map $M_{B}$ is estimated as $\left(1-\max _{1 \leqslant c \leqslant C} M_c\right)$. $M_{B}$ is then concatenated to the initial foreground CAMs, making $M\in \mathbb{R}^{(C+1)\times H\times W}$.
	
	In the next step, $T_{out}$ and $M$ are input to a modified version of the structure-aware seed locating module from SIPE \cite{b23}. This module generates seed maps $R\in \mathbb{R}^{(C+1)\times P\times P}$ for each class category $C+1$, including the background. The module operates by calculating the cross-token semantic affinity map $S_{c}$  from $T_{out}$ to capture each token’s spatial structure. It then compares the similarity between each token’s spatial structure in $S_{c}$ with the class activations in $M$ and assigns that token the class label to which it has the most structural similarity. In contrast to the original method, SWTformer-V2 produces the cross-token semantic similarity map $S_{c}$ by calculating the cosine similarity as 
	\begin{equation}
		S_c(T_{out})=\left| \operatorname{CosSim}(T_{out},T_{out})) \right|= \left|\frac{T_{out} \cdot ({T_{out}})^T}{\left\|T_{out}\right\|\left\|({T_{out}})^T\right\|}\right|,
	\end{equation}
	where  $ \cdot$ denotes the dot product. The use of the absolute value of the similarity is motivated by experiments showing that even a negative value similarity between two tokens represents a high structural correlation between them.
	
	In the final step of this framework, the generated seed maps $R$ and the hierarchical feature $F_{hie}$ are passed to the background-aware prototype modeling module from SIPE \cite{b23}. This module first creates prototypes $P_{c}$ for all (C+1), classes, which are equivalent to the centroid of $R$ for each class in the feature space of $F_{hie}$. Lastly, the refined CAMs, R-CAMs, are generated from the correlation calculated between $P_{c}$ and $F_{hie}$.
	
	To ensure consistency between the initial CAM and the refined R-CAM, the utilization of a normalization loss is suggested by the original paper,
	\begin{equation}
		\mathcal{L}_{\text {GSC }}=\frac{1}{C+1}\left\|C A M-R_{-} C A M\right\|_1.
	\end{equation}	
	SWTformer-V2 proposes to use a Class-wise Contrastive Loss (CCL) in addition to the GSC loss. The CCL loss aims to enhance the generation of comprehensive initial CAMs at each step, building on the R-CAMs generated in the previous step. It achieves this by optimizing the model to minimize the distance between the representations of similar classes and maximize the distance between representations of dissimilar classes, represented in CAM and R-CAM. The CCL loss is calculated as
	\begin{multline}
		\mathcal{L}_{\text {CCL}}=\frac{1}{2}[  (\frac{2}{3} \times CosSim\left(CAM, R_{-} C A M\right))^2 +\\ \left(1-CosSim\left(CAM, R_{-} C A M\right)\right)^2].
	\end{multline}
	In summary, the overall loss for optimizing the training of the model includes the CLS, GSC and CCL loss functions as
	\begin{equation}
		\mathcal{L}_{\text {Total }}= \mathcal{L}_{\text {CLS }} + \mathcal{L}_{\text {GSC }} + \mathcal{L}_{\text {CCL }}
	\end{equation}
	
	\section{Experiments}
	In this section, a detailed explanation of the experimental setup is provided, consisting of the dataset, evaluation metrics and implementation specifics. Subsequently, the performance of the proposed method in object localization and its accuracy in generating initial class activation maps is compared with state-of-the-art methods on the PASCAL VOC 2012 \cite{b33} dataset. Lastly, a set of ablation studies is carried out to confirm the efficacy of the proposed method.
	
	\begin{figure*}[htbp]
		\centering
		\includegraphics[width=\textwidth]{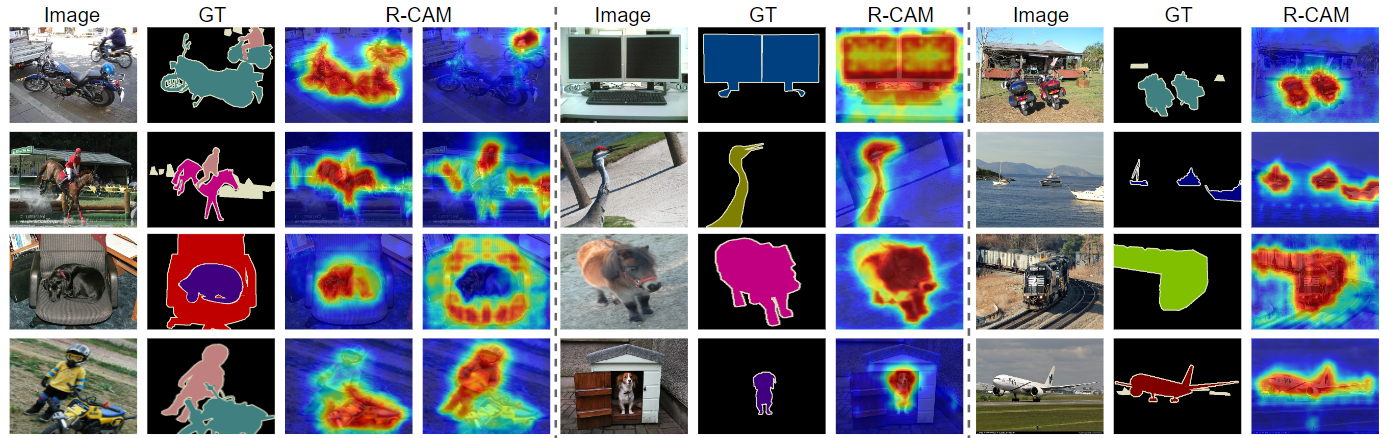}
		\caption{Qualitative results of the class activation maps generated by SWTformer on PASCAL VOC 2012 train set. Images contain singular or multiple class labels.}
		\label{fig_4}
	\end{figure*}
	
	\subsection{Dataset}
	The proposed method is evaluated on the PASCAL VOC 2012 \cite{b33} dataset, a widely used benchmark for image classification and segmentation, 
	particularly for WSSS. The dataset consists of images labeled with 20 foreground and one background classes. 
	Images in PASCAL VOC 2012 are split into 1,464 for training, 1,449 for validation (val), and 1,456 for testing. 
	Following common practice in semantic segmentation, an additional set of 10,582 augmented images from \cite{b34} is also used for training. 
	
	\subsection{Evaluation Metrics}
	The mean Average Precision (mAP) metric is commonly used to evaluate the localization performance of object detection models. In this research, however, it is used to evaluate the localization accuracy of the classifier model. Additionally, the mean Intersection-over-Union (mIoU) metric is employed to compare the accuracy of the generated class activation maps by the proposed method with state-of-the-art approaches.  
	
	\subsection{Implementation Details}
	SWTformer is built using the Swin-T \cite{b7}, pre-trained model on ImageNet \cite{b11} with a $ 224\times 244 $ resolution as the backbone. Images are cropped to $ 224\times 244 $ for training and data augmentation is done 
	following \cite{b35}. The AdamW optimizer is employed to train the model with a batch size of 16 on 2 Nvidia T4 GPUs. Seed maps are equivalent to the refined CAMs, R-CAM.
	
	\subsection{Experimental Results}
	\paragraph{Improvement on object localization} The main objective of this study is to investigate the impact of using the Swin Transformer as the classification backbone for WSSS in localizing objects supervised by image-level labels and generating CAMs. Tab.~\ref{tab_1} compares the localization
	accuracy of the Swin Transformer used in SWTformer with DeiT-S \cite{b10}, which is commonly employed in state-of-the-art WSSS methods using a vision transformer as the backbone. Specifically the localization results of DeiT-S utilized in
	MCTformer \cite{b26} are considered for comparison. The results show that using Swin-T outperforms DeiT-S as a backbone for WSSS by $ 0.98\% $, 
	demonstrating the effectiveness of Swin’s local-to-global view in localizing objects.
	\begin{table}[htpb]
		\centering
		\caption{Comparison of object localization on PASCAL VOC 2012 dataset.  \textsuperscript{\textdagger} denotes our implementation.}
		\label{tab_1}
		\resizebox{\columnwidth}{!}{%
			\begin{tabular}{lcc} 
				\toprule
				\textbf{\textit{Method}}                                                       & \textbf{\textit{Backbone}} & \textbf{\textit{mAP (\%)}}  \\ 
				\toprule
				MCTformer-V1\textsuperscript{\textdagger}\cite{b26} & DeiT-S                     & 95.62                                       \\
				MCTformer-V2\textsuperscript{\textdagger}\cite{b26} & DeiT-S                     & 95.47                                       \\ 
				\midrule
				SWTformer-V1                                                                   & Swin-T                     & 96.49                                       \\
				\textbf{SWTformer-V2}                                                          & \textbf{Swin-T}            & \textbf{96.60}                              \\
				\bottomrule
			\end{tabular}
		}
	\end{table}
	\paragraph{Evaluation of seed localization maps} Given that the generation of seed CAMs is the most crucial step in WSSS, this study aimed to propose a framework for 
	utilizing the Swin Transformer in this process. A comparison of the proposed method with other state-of-the-art methods is presented in Tab.~\ref{tab_2}, demonstrating the mIoU accuracy of the seed maps. To ensure an accurate 
	comparison and evaluate the effectiveness of the Swin Transformer, results from other approaches that rely solely on the backbone are utilized. The comparison reveals that SWTformer-V1 achieves an average of $ 0.82\% $ mIoU accuracy higher than other methods, demonstrating the method’s comparable performance. Furthermore, SWTformer-V2 improves upon SWTformer-V1 by $ 5.32\% $ in mIoU, therefore demonstrating the effectiveness of the strategies proposed to address the limitation of using attention maps from the Swin Transformer for refinement.
	\begin{table}[htpb]
		\centering
		\caption{Evaluation of the initial seed localization maps (Seed) on the PASCAL VOC 2012 train set in terms of mIoU (\%).}
		\label{tab_2}
		\resizebox{\columnwidth}{!}{%
			\begin{tabular}{lccc} 
				\toprule
				\textbf{\textit{Method}} & \textbf{\textit{Year}} & \textbf{\textit{Backbone}} & \textbf{\textit{Seed}}  \\ 
				\toprule
				PSA \cite{b15}                 & 2018                   & VGG-16                     & 48.00                   \\
				IRN \cite{b16}                 & 2019                   & ResNet-50                  & 48.30                   \\
				SEAM \cite{b22}                & 2020                   & ResNet-38                  & 47.43                   \\
				SC-CAM \cite{b24}              & 2020                   & ResNet-101                 & 50.90                   \\
				SIPE \cite{b23}                & 2022                   & ResNet-50                  & 50.10                   \\ 
				\midrule
				MCTformer-V1 \cite{b26}        & 2022                   & DeiT-S                     & 47.20                   \\
				MCTformer-V2 \cite{b26}        & 2022                   & DeiT-S                     & 48.51                   \\
				TransCAM \cite{b29}            & 2023                   & Conformer                  & 51.70                   \\ 
				\midrule
				\textbf{SWTformer-V1}    & \textbf{2024}          & \textbf{Swin-T}            & \textbf{49.84}          \\
				\textbf{SWTformer-V2}    & \textbf{2024}          & \textbf{Swin-T}            & \textbf{55.16}          \\
				\bottomrule
			\end{tabular}
		}
	\end{table}
	\paragraph{Qualitative results} The effectiveness of the proposed approach is further confirmed through various qualitative evaluations of the model's performance. Fig. 4 visualizes refined seed class activation maps (R-CAM) generated by SWTformer for various categories.
	
	\subsection{Ablation Studies}
	The training procedure of the proposed model is optimized by the sum of three loss functions, termed as CLS loss, GSC loss, and CCL loss. An analysis of the impact of each of these loss functions on the enhancement of SWTformer is presented in Tab.~\ref{tab_3}. Experiments demonstrate that the simultaneous use of these three losses results in the best accuracy. Specifically, CLS is responsible for classification, GSC provides consistency between the two sets of CAMs, and CCL further balances these modules.
	\begin{table}[htpb]
		\centering
		\caption{Ablation study on the effectiveness of the proposed loss functions on the accuracy of the seed map.}
		\label{tab_3}
		\resizebox{0.9\columnwidth}{!}{%
			\begin{tabular}{cccc} 
				\toprule
				$ \mathcal{L}_{\text {CLS }} $ & $ \mathcal{L}_{\text {GSC }} $ & $ \mathcal{L}_{\text {CCL }} $ & mIoU (\%)  \\ 
				\toprule
				\checkmark   &     &     & 49.84                      \\
				\checkmark   & \checkmark   &     & 54.58                      \\ 
				\midrule
				\checkmark   & \checkmark   & \checkmark   & \textbf{55.16}             \\
				\bottomrule
			\end{tabular}
		}
	\end{table}
	\section{Conclusion}
	This paper presents a novel approach to utilizing the Swin Transformer as a backbone for WSSS. It introduces SWTformer-V1, which captures both local fine-grained details and global structure by leveraging the hierarchical flow of the transformer. To address the limitations of refining activation maps with the transformer’s attention flow, common in non-hierarchical strategies, SWTformer-V2 is proposed. This version introduces a hierarchical feature fusion
	module to capture multi-scale semantic knowledge from the network, and employs it in a modified version of a background-aware mechanism. SWTformer outperforms state-of-the-art transformers in object localization and also yields comparable results with other approaches in generating seed activation maps. The strategies employed in SWTformer-V2 further enhance this framework by refining the initial activation maps to cover object regions more comprehensively.


\begin{thebibliography}{00}
		\bibitem{b1} E. Goceri and N. Goceri. Deep learning in medical image analysis: recent advances and future trends. In International Association for Development of the Information Society, pp. 305–310, 2017.
		\bibitem{b2} M. Orsic, I. Kreso, P. Bevandic, and S. Segvic. In defense of pre-trained imagenet architectures for real-time semantic segmentation of road-driving images. In Proceedings of the IEEE/CVF Conference on Computer Vision and Pattern Recognition, pp. 12607–12616, 2019.
		\bibitem{b3} J. Dai, K. He, and J. Sun. Boxsup: Exploiting bounding boxes to supervise convolutional networks for semantic segmentation. In Proceedings of the IEEE/CVF International Conference on Computer Vision, pp. 1635–1643, 2015.
		\bibitem{b4} D. Lin, J. Dai, J. Jia, K. He, and J. Sun. Scribblesup: Scribble-supervised convolutional networks for semantic segmentation. In Proceedings of the IEEE/CVF Conference on Computer Vision and Pattern Recognition, pp. 3159–3167, 2016.
		\bibitem{b5} A. Bearman, O. Russakovsky, V. Ferrari, and L. Fei-Fei. What’s the point: Semantic segmentation with point supervision. In Proceedings of the European conference on computer vision,  pp. 549–565, 2016.
		\bibitem{b6} B. Zhou, A. Khosla, A. Lapedriza, A. Oliva, and A. Torralba. Learning deep features for discriminative localization. In Proceedings of the IEEE/CVF Conference on Computer Vision and Pattern Recognition, pp. 2921–2929, 2016.
		\bibitem{b7} Z. Liu, Y. Lin, Y. Cao, H. Hu, Y. Wei, Z. Zhang, S. Lin, and B. Guo. Swin transformer: Hierarchical vision transformer using shifted windows. In Proceedings of the IEEE/CVF International Conference on Computer Vision, pp. 10012–10022, 2021.
		\bibitem{b8} S. Abnar and W. Zuidema. Quantifying attention flow in transformers. In arXiv preprint arXiv:2005.00928, 2020.
		\bibitem{b9} A. Dosovitskiy, L. Beyer, A. Kolesnikov, D. Weissenborn, X. Zhai, T. Unterthiner, M. Dehghani, M. Minderer, G. Heigold, S. Gelly, J. Uszkoreit, and N. Houlsby. An image is worth 16x16 words: Transformers for image recognition at scale. In International Conference on Learning Representations, 2021.
		\bibitem{b10} H. Touvron, M. Cord, M. Douze, F. Massa, A. Sablayrolles, and H. Jégou. Training data-efficient image transformers \& distillation through attention. In International conference on machine learning, pp. 10347–10357, 2021.
		\bibitem{b11} J. Deng, W. Dong, R. Socher, L.-J. Li, K. Li, and L. Fei-Fei. Imagenet: A largescale hierarchical image database. In Proceedings of the IEEE/CVF Conference on Computer Vision and Pattern Recognition, pp. 248–255, 2009.
		\bibitem{b12} A. Peng, Zhiliang and Huang, Wei and Gu, Shanzhi and Xie, Lingxi and Wang, Yaowei and Jiao, Jianbin and Ye, Qixiang. Conformer: Local features coupling global representations for visual recognition. In Proceedings of the IEEE/CVF International Conference on Computer Vision, pp. 367-376, 2021.
		\bibitem{b13} L. Yuan, Y. Chen, T. Wang, W. Yu, Y. Shi, Z.-H. Jiang, F. E. Tay, J. Feng, and S. Yan. Tokens-to-token vit: Training vision transformers from scratch on imagenet. In Proceedings of the IEEE/CVF International Conference on Computer Vision, pp. 558–567, 2021.
		\bibitem{b14} W. Wang, E. Xie, X. Li, D.-P. Fan, K. Song, D. Liang, T. Lu, P. Luo, and L. Shao. Pyramid vision transformer: A versatile backbone for dense prediction without convolutions. In Proceedings of the IEEE/CVF International Conference on Computer Vision, pp. 568–578, 2021.
		\bibitem{b15} J. Ahn and S. Kwak. Learning pixel-level semantic affinity with image-level supervision for weakly supervised semantic segmentation. In Proceedings of the IEEE/CVF Conference on Computer Vision and Pattern Recognition, pp. 4981–4990, 2018.
		\bibitem{b16} J. Ahn, S. Cho, and S. Kwak. Weakly supervised learning of instance segmentation with inter-pixel relations. In Proceedings of the IEEE/CVF Conference on Computer Vision and Pattern Recognition, pp. 2209–2218, 2019.
		\bibitem{b17} Y. Wei, J. Feng, X. Liang, M.-M. Cheng, Y. Zhao, and S. Yan. Object region mining with adversarial erasing: A simple classification to semantic segmentation approach. In Proceedings of the IEEE/CVF Conference on Computer Vision and Pattern Recognition, volume 31, pp. 1568–1576, 2017.
		\bibitem{b18} J. Lee, E. Kim, S. Lee, J. Lee, and S. Yoon. Ficklenet: Weakly and semi-supervised semantic image segmentation using stochastic inference. In Proceedings of the IEEE/CVF Conference on Computer Vision and Pattern Recognition, pp. 5267–5276, 2019.
		\bibitem{b19} T. Zhou, M. Zhang, F. Zhao, and J. Li. Regional semantic contrast and aggregation for weakly supervised semantic segmentation. In Proceedings of the IEEE/CVF Conference on Computer Vision and Pattern Recognition, pp. 4299– 4309, 2022.
		\bibitem{b20} Y. Du, Z. Fu, Q. Liu, and Y. Wang. Weakly supervised semantic segmentation by pixel-to-prototype contrast. In Proceedings of the IEEE/CVF Conference on Computer Vision and Pattern Recognition, pp. 4320–4329, 2022.
		\bibitem{b21} T. Wu, J. Huang, G. Gao, X. Wei, X. Wei, X. Luo, and C. H. Liu. Embedded discriminative attention mechanism for weakly supervised semantic segmentation. In Proceedings of the IEEE/CVF Conference on Computer Vision and Pattern Recognition , pp. 16765–16774, 2021.
		\bibitem{b22} Y. Wang, J. Zhang, M. Kan, S. Shan, and X. Chen. Self-supervised equivariant attention mechanism for weakly supervised semantic segmentation. In Proceedings of the IEEE/CVF Conference on Computer Vision and Pattern Recognition, pp. 12275–12284, 2020.
		\bibitem{b23} Q. Chen, L. Yang, J.-H. Lai, and X. Xie. Self-supervised image-specific prototype exploration for weakly supervised semantic segmentation. In Proceedings of the IEEE/CVF Conference on Computer Vision and Pattern Recognition, pp. 4288–4298, 2022.
		\bibitem{b24} Y.-T. Chang, Q. Wang, W.-C. Hung, R. Piramuthu, Y.-H. Tsai, and M.-H. Yang. Weakly-supervised semantic segmentation via sub-category exploration. In Proceedings of the IEEE/CVF Conference on Computer Vision and Pattern Recognition, pp. 8991–9000, 2020.
		\bibitem{b25} L. Ru, Y. Zhan, B. Yu, and B. Du. Learning affinity from attention: End-to-end weakly-supervised semantic segmentation with transformers. In Proceedings of the IEEE/CVF Conference on Computer Vision and Pattern Recognition, pp. 16846–16855, 2022.
		\bibitem{b26} L. Xu, W. Ouyang, M. Bennamoun, F. Boussaid, and D. Xu. Multi-class token transformer for weakly supervised semantic segmentation. In Proceedings of the IEEE/CVF Conference on Computer Vision and Pattern Recognition, pp. 4310–4319, 2022.
		\bibitem{b27} S. Rossetti, D. Zappia, M. Sanzari, M. Schaerf, and F. Pirri. Max pooling with vision transformers reconciles class and shape in weakly supervised semantic segmentation. In Proceedings of the European conference on computer vision, pp. 446–463, 2022.
		\bibitem{b28} L. Ru, H. Zheng, Y. Zhan, and B. Du. Token contrast for weakly-supervised semantic segmentation. In Proceedings of the IEEE/CVF Conference on Computer Vision and Pattern Recognition, pp. 3093–3102, 2023.
		\bibitem{b29} R. Li, Z. Mai, Z. Zhang, J. Jang, and S. Sanner. Transcam: Transformer attentionbased cam refinement for weakly supervised semantic segmentation. In Elsevier Journal of Visual Communication and Image Representation, volume 92, pp. 103800, 2023.
		\bibitem{b30} L. Xu, W. Ouyang, M. Bennamoun, F. Boussaid, and D. Xu. Multi-class token transformer for weakly supervised semantic segmentation. In Proceedings of the IEEE/CVF Conference on Computer Vision and Pattern Recognition, pp. 4310–4319, 2022.
		\bibitem{b31} T.-Y. Lin, P. Dollár, R. Girshick, K. He, B. Hariharan, and S. Belongie. Feature pyramid networks for object detection. In Proceedings of the IEEE/CVF Conference on Computer Vision and Pattern Recognition, pp. 2117–2125, 2017.
		\bibitem{b32} T. Chen and L. Mo. Swin-fusion: swin-transformer with feature fusion for human action recognition. In Springer Neural Processing Letters, pp. 11109–11130, 2023.
		\bibitem{b33} M. Everingham, L. Van Gool, C. K. Williams, J. Winn, and A. Zisserman. The pascal visual object classes (voc) challenge. In Springer International journal of computer vision, volume 88, pp. 303–338, 2010.
		\bibitem{b34} B. Hariharan, P. Arbeláez, L. Bourdev, S. Maji, and J. Malik. Semantic contours from inverse detectors. In Proceedings of the IEEE/CVF International Conference on Computer Vision, volume 90, pp. 991–998, 2011.
		\bibitem{b35} J. Lee, E. Kim, and S. Yoon. Anti-adversarially manipulated attributions for weakly and semi-supervised semantic segmentation. In Proceedings of the IEEE/CVF Conference on Computer Vision and Pattern Recognition, pp. 4071–4080, 2021.
		
	\end{thebibliography}
\end{document}